\documentclass[final]{cvpr}

\usepackage{times}
\usepackage{epsfig}
\usepackage{graphicx}
\usepackage{amsmath}
\usepackage{amssymb}

\usepackage{comment}
\usepackage{amsmath,amssymb}
\usepackage{algorithm}
\usepackage{algorithmic}
\usepackage{tabularx}
\usepackage{multirow}
\usepackage{bm}
\usepackage{booktabs}
\usepackage{array}
\usepackage{microtype}
\usepackage{arydshln}
\usepackage[switch]{lineno}

\usepackage{pifont}
\usepackage[usenames,dvipsnames]{xcolor}
\definecolor{dgreen}{rgb}{0.0,0.6,0.0}
\definecolor{dred}{rgb}{0.6,0.0,0.0}
\definecolor{alexey}{rgb}{0.7,0,1}
\definecolor{philipcolor}{rgb}{0,0.5,0}
\definecolor{grey}{rgb}{0.6,0.6,0.6}
\definecolor{dblue}{rgb}{0.0,0.0,0.7}

\newcommand{\cmark}{\textcolor{dgreen}{\text{\ding{51}}}}
\newcommand{\xmark}{\textcolor{dred}{\text{\ding{55}}}}

\newcommand{\adaresnetcorr}{\textit{Ada-ResNetCorr}}
\newcommand{\adapsmnet}{\textit{Ada-PSMNet}}

\usepackage{soul}

\usepackage[pagebackref=true,breaklinks=true,colorlinks,bookmarks=false]{hyperref}

\newcommand\nnfootnote[1]{
  \begin{NoHyper}
  \renewcommand\thefootnote{}\footnote{#1}
  \addtocounter{footnote}{-1}
  \end{NoHyper}
}

\def\cvprPaperID{4004}
\def\confYear{CVPR 2021}

\begin{document}

\title{AdaStereo: A Simple and Efficient Approach for Adaptive Stereo Matching}

\author{
   {Xiao Song}${^{1\star}}$ \quad
   {Guorun Yang}${^{1,3\star}}$ \quad
   {Xinge Zhu}${^{2}}$ \quad
   {Hui Zhou}${^{1}}$ \quad
   {Zhe Wang}${^{1,4}}$ \quad
   {Jianping Shi}${^{1,5}}$\\
   ${^1}${SenseTime Research} \quad
   ${^2}${The Chinese University of Hong Kong}\\
   ${^3}${Shenzhen Institutes of Advanced Technology, Chinese Academy of Sciences} \\
   ${^4}${Shanghai AI Laboratory} \quad
   ${^5}${Qing Yuan Research Institute, Shanghai Jiao Tong University}\\
   {\tt\small \{songxiao,yangguorun,zhouhui,wangzhe,shijianping\}@sensetime.com \, zx018@ie.cuhk.edu.hk}
}

\maketitle

\nnfootnote{$^{\star}$ indicates equal contribution.}

\pagestyle{empty}
\thispagestyle{empty}

	\begin{abstract}
	
	
	
	 Recently, records on stereo matching benchmarks are constantly broken by end-to-end disparity networks. However, the domain adaptation ability of these deep models is quite poor. Addressing such problem, we present a novel domain-adaptive pipeline called AdaStereo that aims to align multi-level representations for deep stereo matching networks. Compared to previous methods for adaptive stereo matching, our AdaStereo realizes a more standard, complete and effective domain adaptation pipeline. Firstly, we propose a non-adversarial progressive color transfer algorithm for input image-level alignment. Secondly, we design an efficient parameter-free cost normalization layer for internal feature-level alignment. Lastly, a highly related auxiliary task, self-supervised occlusion-aware reconstruction is presented to narrow down the gaps in output space. Our AdaStereo models achieve state-of-the-art cross-domain performance on multiple stereo benchmarks, including KITTI, Middlebury, ETH3D, and DrivingStereo, even outperforming disparity networks finetuned with target-domain ground-truths.
	\end{abstract}

\section{Introduction}
\label{sec:intro}
	
	The stereo matching task aims to find all corresponding pixels in a stereo pair, and the distance between corresponding pixels is known as disparity~\cite{hartley2003multiple}. Based on the epipolar geometry, stereo matching enables stable depth perception, hence it supports further applications such as scene understanding, object detection, odometry, and SLAM.
	
\begin{figure}[!t]
		\centering
		\includegraphics[width=0.95\linewidth]{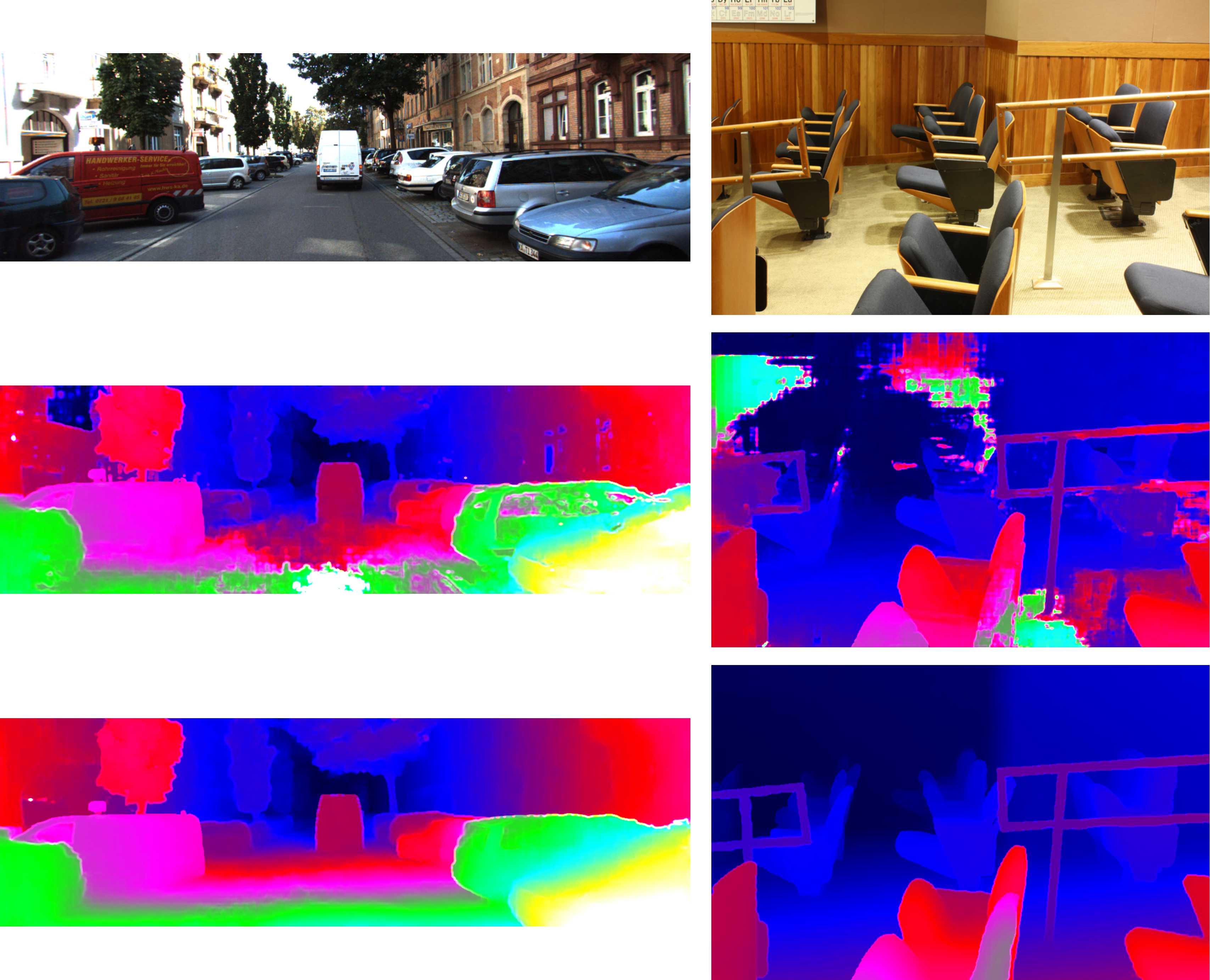}
		\caption{
			Overview examples. Left-right: KITTI~\cite{Menze2015CVPR} and Middlebury~\cite{Scharstein2014High}. Top-down: left image, disparity maps predicted by the SceneFlow-pretrained PSMNet~\cite{chang2018pyramid}, and by our Ada-PSMNet.
		}
		\label{fig:overview_examples}
	\end{figure}

	Recent stereo matching methods typically adopt fully convolutional networks~\cite{long2015fully} to regress disparity maps directly and have achieved state-of-the-art performance on stereo benchmarks~\cite{Geiger2012CVPR,Menze2015CVPR,Scharstein2014High}. However, the performance of these methods adapted from synthetic data to real-world scenes is  limited. As shown in Fig.~\ref{fig:overview_examples}, the PSMNet~\cite{chang2018pyramid} pretrained on the SceneFlow dataset~\cite{mayer2016large} fails to produce good results on the Middlebury~\cite{Scharstein2014High} and KITTI~\cite{Menze2015CVPR} datasets. Therefore, instead of designing complicated networks for higher accuracy on specific datasets, how to obtain effective domain-adaptive stereo matching models is more desirable now.

\begin{figure}[tb]
		\centering
		\includegraphics[width=1.0\linewidth]{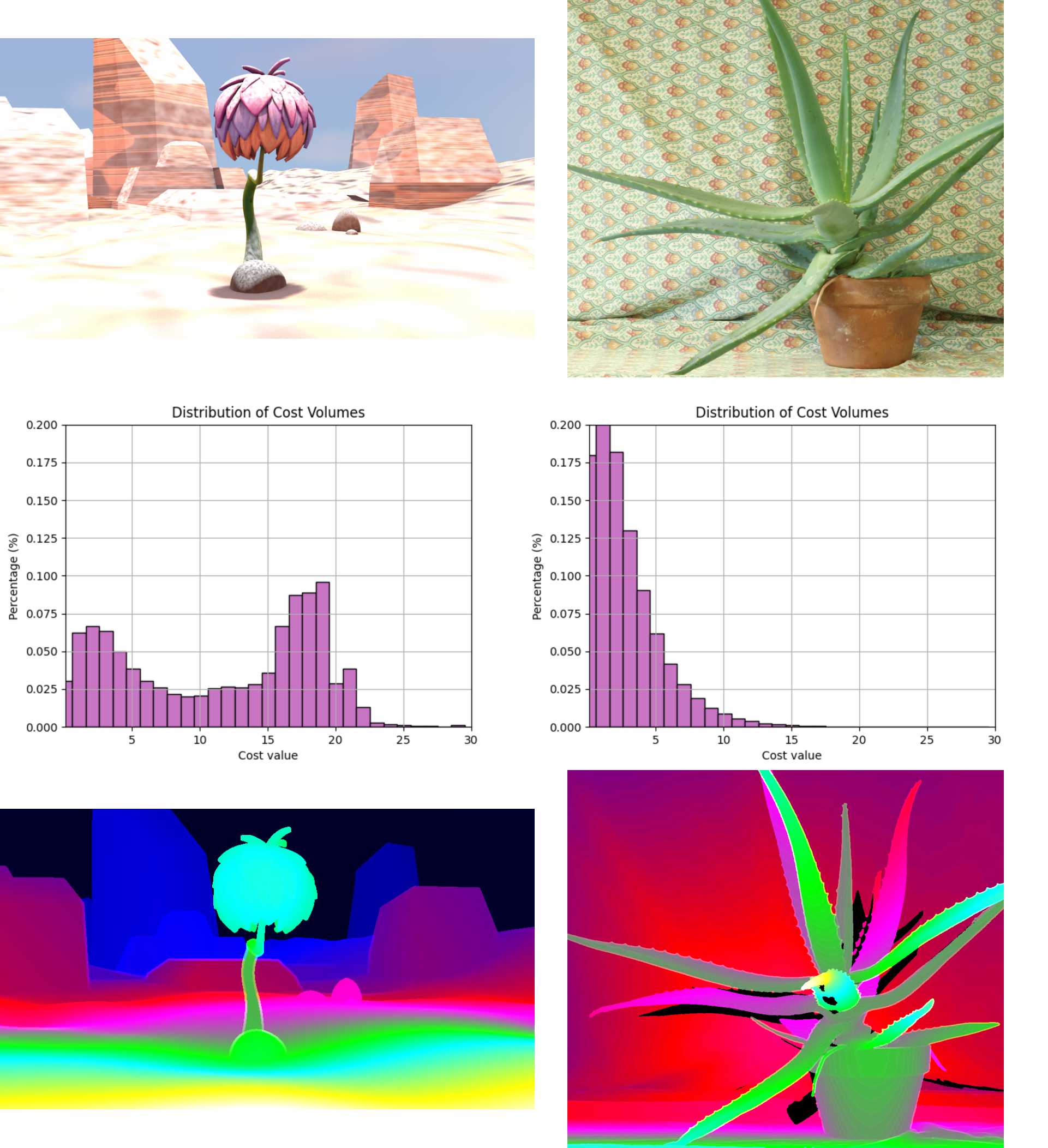}
		\caption{
			Comparisons of input images, internal cost volumes, and output disparity  maps between synthetic and real-world datasets. Left-right: SceneFlow~\cite{mayer2016large} and Middlebury~\cite{Scharstein2014High}. Top-down: input image, internal cost volume, and output disparity map. Disparity maps are rendered by the same color map.
		}
		\label{fig:feature_visualization}
	\end{figure}

    In this work, we aim at the important but less explored problem of domain adaptation in stereo matching. Considering the fact that there are a great number of synthetic data but only a small number of realistic data with ground-truths, we focus on domain gaps between synthetic and realistic domains. We first analyse main differences between these two domains, as shown in Fig.~\ref{fig:feature_visualization}. At the input image level, color and brightness are the obvious gaps. By making statistics on the internal cost volumes, we also find significant differences in distributions. Moreover, geometries of the output disparity maps are inconsistent as well. In order to bridge the domain gaps at these levels (input image, internal cost volume, and output disparity), we propose \emph{\textbf{AdaStereo}}, a standard and complete domain adaptation pipeline for stereo matching, in which three particular modules are presented:

	
	
	\begin{itemize}
	    \item For input image-level alignment, the \textbf{non-adversarial progressive color transfer} algorithm is presented to align input color space between source and target domains during training. It is the first attempt that adopts a non-adversarial style transfer method to align input-level inconsistency for stereo domain adaptation, avoiding harmful side-effects of geometrical distortions from GAN-based methods \cite{zhu2017unpaired}. Furthermore, the proposed progressive update strategy enables capturing representative target-domain color styles during adaptation.
	
	
	
	    \item For internal feature-level alignment, the \textbf{cost normalization} layer is proposed to align matching cost distribution. Oriented to the stereo matching task, two normalization operations are designed and embedded in this layer: (i) channel normalization reduces the inconsistency in scaling of each feature channel; and (ii) pixel normalization further regulates the norm distribution of pixel-wise feature vector for binocular matching. Compared to previous general normalization layers (\emph{e.g.} IN \cite{ulyanov2016instance}, DN \cite{zhang2020domain}), our cost normalization layer is parameter-free and adopted only once in the network.
	
	

	
	
	    \item For output-space alignment, we conduct self-supervised learning through a highly related auxiliary task, \textbf{self-supervised occlusion-aware reconstruction}, which is the first proposed auxiliary task for stereo domain adaptation. Concretely, a self-supervised module is attached upon the main disparity network, to perform image reconstructions on the target domain. To address the ill-posed occlusion problem in reconstruction, we also design a domain-collaborative learning strategy for occlusion mask predictions. Through occlusion-aware stereo reconstruction, more informative geometries from target scenes are involved in model training, thus benefiting disparity predictions across domains.
	
	
    \end{itemize}

	Based on our proposed pipeline, we conduct effective domain adaptation from synthetic data to real-world scenes. In Fig.~\ref{fig:overview_examples}, our Ada-PSMNet pretrained on the synthetic dataset performs well on both indoor and outdoor scenes. In order to validate the effectiveness of each module, ablation studies are performed on diverse real-world datasets, including Middlebury~\cite{Scharstein2014High}, ETH3D~\cite{schoeps2017cvpr}, KITTI~\cite{Geiger2012CVPR,Menze2015CVPR}, and DrivingStereo~\cite{yang2019drivingstereo}. Our domain-adaptive models outperform other traditional / domain generalization / domain adaptation methods, and even finetuned models on multiple stereo matching benchmarks. Main contributions are summarized below:
	
	\begin{itemize}
		\item
		We locate the domain-adaptive problem and investigate domain gaps for deep stereo matching networks.
		\item
		We propose a novel domain adaptation pipeline, including three modules to narrow down the gaps at input image level, internal feature level, and output space.
		\item
		Our domain-adaptive models outperform other domain-invariant methods, and even finetuned disparity networks on multiple stereo matching benchmarks.
		
		
	\end{itemize}
	
\section{Related Work}
	\label{sec:related_work}
	
	\subsection{Stereo Matching}
	

    Recently, end-to-end stereo matching networks have achieved state-of-the-art performance, which can be roughly categorized into two types: correlation-based $2$-D stereo networks and cost-volume based $3$-D stereo networks. On the one hand, Mayer \emph{et al.} \cite{mayer2016large} proposed the first end-to-end disparity network DispNetC. Since then, based on color or feature correlations, more advanced models were proposed, including CRL \cite{pang2017cascade}, iResNet \cite{liang2017learning}, HD$^{3}$ \cite{yin2019hierarchical}, SegStereo \cite{yang2018SegStereo}, and EdgeStereo \cite{song2018stereo,song2020edgestereo}. On the other hand, $3$-D convolutional neural networks show the advantages in regularizing cost volume for disparity estimation, including GC-Net \cite{kendall2017end}, PSMNet \cite{chang2018pyramid}, GWCNet \cite{guo2019group}, GANet \cite{zhang2019ga}, \emph{etc.} Our proposed domain adaptation pipeline for stereo matching can be easily applied on both $2$-D and $3$-D stereo networks.


	\subsection{Domain Adaptation}
	
     Prior works on domain adaptation can be roughly divided into two categories. The general idea of the first category is aligning source and target domains at different levels, including: (1) input image-level alignment \cite{bousmalis2017unsupervised,hoffman2018cycada}, using image-to-image translation methods such as CycleGAN \cite{zhu2017unpaired}; (2) internal feature-level alignment,  based on feature-level domain adversarial learning \cite{tzeng2017adversarial,long2018conditional}; (3) conventional discrepancy measures,  such as MMD \cite{MMD} and CMD \cite{zellinger2017central}; and (4) output-space alignment \cite{adapt,vu2019advent}, based on adversarial learning. For the second category, self-supervised learning based domain adaptation methods \cite{sun2019unsupervised} achieve great progress, in which simple auxiliary tasks generated automatically from unlabeled data are utilized to train feature representations, such as rotation prediction \cite{gidaris2018unsupervised}, patch-location prediction \cite{xu2019self},~\etc. In this paper, we explicitly implement domain alignments at input level and internal feature level, while incorporating self-supervised learning into output-space alignment through a specifically designed auxiliary task.

	
   \subsection{Domain-Adaptive Stereo Matching}



     Although records on public benchmarks are constantly broken, few attention has been paid to the domain adaptation ability of deep stereo models. Pang \emph{et al.} \cite{pang2018zoom} proposed a semi-supervised method utilizing the scale information. Guo \emph{et al.} \cite{guo2018learning} presented a cross-domain method using knowledge distillation. MAD-Net \cite{tonioni2019real} was designed to adapt a compact stereo model online. Recently, StereoGAN~\cite{liu2020stereogan} utilized CycleGAN~\cite{zhu2017unpaired} to bridge domain gaps by joint optimizations of image style transfer and stereo matching. However, no standard and complete domain adaptation pipeline was implemented in these methods, and their adaptation performance is quite limited. Contrarily, we propose a more complete pipeline for deep stereo models following the standard domain adaptation methodology, in which alignments across domains are conducted at multiple levels thereby remarkable adaptation performance is achieved. In addition, we do not conduct any adversarial learning, hence the training stability and semantic invariance are guaranteed.


\section{Method}
	\label{sec:method}


    In this section, we first describe the problem of domain-adaptive stereo matching. Then we introduce the motivation and give an overview of our domain adaptation pipeline. After that, we detail the main components in the pipeline, \ie non-adversarial progressive color transfer, cost normalization, and self-supervised occlusion-aware reconstruction.
	
	\subsection{Problem Description}
	\label{sec:formulation}
	
	In this paper, we focus on the domain adaptation problem for stereo matching. Different from domain generalization where a method needs to perform well on unseen scenes, domain adaptation allows using target-domain images without ground-truths during training. Specifically for stereo matching, since there are a great number of synthetic data~\cite{mayer2016large} but only a small number of realistic data with ground-truths~\cite{Menze2015CVPR,Scharstein2014High,schoeps2017cvpr}, the problem can be further limited to the adaptation from virtual to real-world scenarios. Given stereo image pairs $({I_s^l}, {I_s^r})$ and $({I_t^l}, {I_t^r})$ on source synthetic and target realistic domains, and the ground-truth disparity map $\hat{d_{s}^{l}}$ on the source synthetic domain, we train the model to predict the disparity map $d_t^l$ on the target domain.
	
	\subsection{Motivation}
	\label{sec:motivation}
		

	As shown in Fig.~\ref{fig:feature_visualization}, two images from the  SceneFlow~\cite{mayer2016large} and Middlebury~\cite{Scharstein2014High} datasets are selected to describe inherent inconsistencies between two domains. (i) These two images own observable differences in their color and brightness. Moreover, according to the statistics on the \emph{whole datasets}, the mean values of RGB channels are $(107, 102, 92)$ in SceneFlow and $(148, 132, 102)$ in Middlebury. Therefore, significant color variances are found between synthetic and realistic domains. (ii) For cost volumes computed from the 1D-correlation layer ~\cite{mayer2016large}, we calculate the proportion of matching cost values in each interval and find the distributions between two domains are inconsistent as well. (iii) Although these two images have similar plants as foreground, the generated disparity maps still vary in scene geometries. Both the foreground objects and background screens have quite different disparities. Therefore, we conclude that the inherent differences across domains for stereo matching lie in the image color at input level, cost volume at feature level, and disparity map at output level.
	
	
	
	Correspondingly, to solve the domain adaptation problem for stereo matching, we propose the progressive color transfer, cost normalization, and self-supervised reconstruction to handle the domain gaps in three levels respectively. The former two methods are presented to narrow down the differences in color space and matching cost distribution  directly. The latter reconstruction is appended as an auxiliary task to impose extra supervision on disparity regression for target domain, finally benefiting the domain  adaptation ability.
	
	
	\begin{figure*}[!t]
	\centering
	\includegraphics[width=0.99\textwidth]{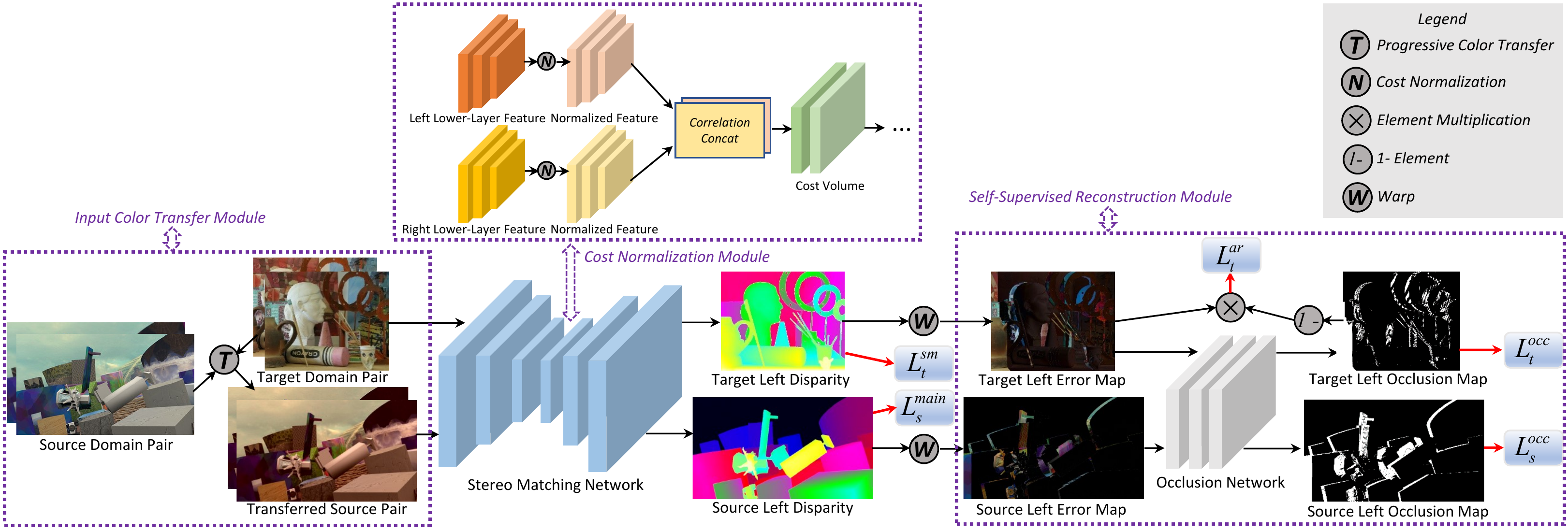}
	\caption{
		The training diagram of AdaStereo, with the adaptation from SceneFlow to Middlebury as an example. Color transfer and self-supervised occlusion-aware reconstruction modules are only adopted during training. ($L_{s}^{main}, L_{s}^{occ}$, $L_{t}^{ar}$, $L_{t}^{occ}$,  $L_{t}^{sm}$) are the five training loss terms specified in Eq.~\ref{loss_t}.
	}
	\label{fig:training_diagram}
	\end{figure*}
	
	\subsection{Framework Overview}
	

    Fig.~\ref{fig:training_diagram} depicts the training pipeline of our stereo domain adaptation framework, in which three proposed modules are involved. For input, a randomly selected target-domain pair, and a randomly selected source-domain pair which adapts to target-domain color styles through our progressive color transfer algorithm, are simultaneously fed into a shared-weight disparity network with our cost normalization layer. The source-domain branch is under the supervision of the  given ground-truth disparity map, while the target-domain branch is regulated by the proposed auxiliary task: self-supervised occlusion-aware reconstruction.

	
	

	\subsection{Non-adversarial Progressive Color Transfer}
	
	
	
	As mentioned in Sec.~\ref{sec:motivation}, color difference plays a major role in the input image-level inconsistency across domains. Hence, we present an effective and stable algorithm for color transfer from source domain to target domain in a non-adversarial manner. During training, given a source-domain image $I_s$ and a target-domain image $I_t$, the algorithm outputs a transferred source-domain image $I_{s\rightarrow t}$, which preserves the contents of $I_s$ and owns the target-domain color styles.  As in Alg.~\ref{alg:color_transfer}, color transfer is performed in the $LAB$ color space. ${T}_{RGB{\rightarrow}LAB}(.)$ and ${T}_{LAB{\rightarrow}RGB}(.)$ denote the color space transformations. Under the $LAB$ space, the mean value $\mu$ and standard deviation $\sigma$ of each channel are first computed by $S(\cdot)$. Then, each channel in the source-domain $LAB$ image $\tilde{I_s}$ is subtracted by its mean ${\mu}_{s}$ and multiplied by the standard deviation ratio $\lambda$. Finally, the transferred image $I_{s\rightarrow t}$ is obtained through the addition of the progressively updated ${\mu}_{t}$ and color space conversion. During training, two images in a source-domain pair are simultaneously transferred with the same ${\mu}_{t}$ and $ {\sigma}_{t}$.
	
	Compared with the Reinhard's color transfer method~\cite{reinhard2001color}, the main contribution of our algorithm is the proposed progressive update strategy that proves to be more beneficial for domain adaptation. Considering color inconsistencies might exist across different images in the same target-domain dataset while the previous method ~\cite{reinhard2001color} only allows one-to-one transformations, the  source-domain images can not adapt to meaningful color styles that are representative for the whole target-domain dataset during adaptation. The progressive update  strategy is proposed to address such problem. To be specific, target-domain ${\mu}_{t}$ and ${\sigma}_{t}$ are progressively re-weighted by current inputs $({{\mu}_{t}}^{i}, {{\sigma}_{t}}^{i} )$ and historical records $({\mu}_{t}, {\sigma}_{t})$ with a momentum $\gamma$, simultaneously ensuring the diversity and representativeness of target-domain color styles during adaptation. Experimental results further validate its effectiveness over the previous color transfer method ~\cite{reinhard2001color}.

	In a larger sense, we are the first to use a non-adversarial style transfer method to align input-level inconsistency for stereo domain adaptation. Unlike GAN-based style/color transfer networks~\cite{li2017watergan,zhu2017unpaired} that cause harmful  side-effects of geometrical distortions for the low-level stereo matching task, our algorithm is more stable and training-efficient, which can be easily embedded in the training framework of stereo domain adaptation. Experimental results further validate its superiority over other adversarial transfer methods.

	\begin{algorithm}
		\caption{\textbf{Progressive Color Transfer}}
		\label{alg:color_transfer}
		\begin{algorithmic}[1]
			\REQUIRE {Source-domain dataset $D_s$, target-domain dataset $D_t$, ${\mu}_{t}=0$, ${\sigma}_{t}=0$}
			\STATE Randomly shuffle $D_s$ and $D_t$
			\FOR{$i \in [0, len(D_s))$}
			\STATE Select $I_s \in D_s$, $I_t \in D_t$
			\STATE $\tilde{I_s}{~\Leftarrow~}{T}_{RGB{\rightarrow}LAB}(I_s)$, \space\space $\tilde{I_t}{~\Leftarrow~}{T}_{RGB{\rightarrow}LAB}(I_t)$
			\STATE $({\mu}_s, {\sigma}_s){~\Leftarrow~}{S{(\tilde{I_s})}}$,
			\space\space
			$({{\mu}_{t}}^{i}, {{\sigma}_{t}}^{i}){~\Leftarrow~}{S{(\tilde{I_t})}}$
			\STATE
			${\mu}_{t}{~\Leftarrow~} (1-\gamma)*{\mu}_{t}+ \gamma*{{\mu}_{t}}^{i}$
			\STATE
			$ {\sigma}_{t}{~\Leftarrow~}{(1-\gamma)*{\sigma}_{t}+\gamma*{{\sigma}_{t}}^{i}}$
			\STATE
			$\tilde{I_s}{~\Leftarrow~}{\tilde{I_s}-{\mu}_s} $, ${\lambda}{~\Leftarrow~}{{\sigma}_t / {\sigma}_s}$
			\STATE $\tilde{I_{s\rightarrow t}}{~\Leftarrow~}{\lambda * \tilde{I_s}+ {\mu}_t}$
			\STATE ${I_{s\rightarrow t}}{~\Leftarrow~}{T}_{LAB{\rightarrow}RGB}(\tilde{I_{s\rightarrow t}}) $
			\ENDFOR
		\end{algorithmic}
	\end{algorithm}
	
	\subsection{Cost Normalization}

    Cost volume is the most important internal feature-level representation in a deep stereo network, encoding all necessary information for succeeding disparity regression.
    Hence for domain-adaptive stereo matching, an intuitive way is to directly narrow down the deviations in matching cost distributions across domains.
    Correspondingly, we design the cost normalization layer, which is compatible with all cost volume building patterns (correlation and concatenation) in stereo matching, as shown in Fig.~\ref{fig:training_diagram}.

	
    Before cost volume construction, the left lower-layer feature $\mathcal{F^{L}}$ and right feature $\mathcal{F^{R}}$ with the same size of $N \times C \times H \times W$ ($N$: batch size, $C$: channel, $H$: spatial height, $W$: spatial width), are both successively regularized by two proposed normalization operations: channel normalization and pixel normalization. Specifically, the channel normalization is applied across all spatial dimensions ($H, W$) per channel per sample
    individually, which is defined as:
    \begin{equation}
    \mathcal{F}_{n,c,h,w} = \frac{\mathcal{F}_{n,c,h,w}}{\sqrt{\sum_{h=0}^{H-1}{\sum_{w=0}^{W-1}{||\mathcal{F}_{n,c,h,w}||^{2}}}+\varepsilon}},
    \end{equation}
    where $\mathcal{F}$ denotes the lower-layer feature, ${h}$ and ${w}$ denote the spatial position, $c$ denotes the channel, and $n$ denotes the batch index. After the channel normalization, the pixel normalization is further applied across all channels per spatial position per sample individually, which is defined as:
    \begin{equation}
    \mathcal{F}_{n,c,h,w} = \frac{\mathcal{F}_{n,c,h,w}}{\sqrt{\sum_{c=0}^{C-1}{||\mathcal{F}_{n,c,h,w}||^{2}}+\varepsilon}}.
    \end{equation}
    Through channel normalization which reduces the inconsistency in norm and scaling of each feature channel, and pixel normalization which further regulates the norm distribution of pixel-wise feature vector for binocular matching, inter-domain gaps in matching cost distributions due to varied image contents and geometries are greatly reduced.

    In a nutshell, our parameter-free cost normalization layer is indeed a normalization layer designed specifically for stereo domain adaptation, which is adopted only once before cost volume construction. On the contrary, previous normalization layers (\emph{e.g.} BIN \cite{nam2018batch}, IN \cite{ulyanov2016instance}, CN \cite{dai2019channel} (just IN), and DN \cite{zhang2020domain}) are general normalization approaches, which contain learnable parameters and are repeatedly adopted in the network's feature extractor. Hence regulations on cost volume from these general normalization layers are not direct and effective enough, requiring extra implicit learning process. Moreover, our cost normalization layer does not use zero-centralization to prevent extra disturbances in matching cost distributions. Experimental results further validate its superiority over other general normalization layers.
    	


    \subsection{\textbf{\small{Self-Supervised Occlusion-Aware Reconstruction}}}


    Self-supervised auxiliary tasks are demonstrated to be beneficial for domain adaptation  on high-level tasks \cite{gidaris2018unsupervised,xu2019self}. However, such methodology has not been explored for the low-level stereo matching task. In this subsection, we propose an effective auxiliary task for stereo domain adaptation: self-supervised occlusion-aware reconstruction. As shown in Fig.~\ref{fig:training_diagram}, a self-supervised module is attached upon the main disparity network, to perform image reconstructions on the target domain. To address the ill-posed occlusion problem in reconstruction, we design a domain-collaborative learning strategy for occlusion mask predictions. Through occlusion-aware stereo reconstruction, more informative geometries from target scenes are involved in training.

    During the self-supervised learning, stereo reconstruction is firstly measured by  differences between the input target-domain left image $I_{t}^{l}$ and the corresponding warped image $\overline{I_{t}^{l}}$ (based on the right image $I_{t}^{r}$ and the produced disparity map $d_{t}^{l}$). Then, a small fully-convolutional occlusion prediction network takes the concatenation of $d_{t}^{l}$, $I_{t}^{r}$, and the pixel-wise error map $e_{t}^{l}=|I_{t}^{l}-\overline{I_{t}^{l}}|$ as input, and produces a pixel-wise occlusion mask $O_{t}^{l}$ whose element denotes per-pixel occlusion probability from $0$ to $1$. Next, the reconstruction loss $L_{t}^{ar}$ is re-weighted by the occlusion mask $O_{t}^{l}$ and error map $e_{t}^{l}$ on each pixel. Furthermore, we introduce the disparity smoothness loss ($L_{t}^{sm}$) to avoid possible artifacts. To guide the occlusion mask learning on the target domain more effectively, the shared-weight occlusion prediction network simultaneously learns an occlusion mask $O_{s}^{l}$ on the source domain, under the supervision of the ground-truth occlusion mask $\hat{O_{s}^{l}}$ generated from the ground-truth disparity map $\hat{d_{s}^{l}}$. More details are provided in the supplementary material.

    Our self-supervised occlusion-aware reconstruction task is the first proposed auxiliary task for stereo domain adaptation. In addition, our design enables collaborative occlusion mask learning on both source and target domains, acting as another domain adaptation on occlusion prediction that ensures the quality of the target-domain occlusion mask and explicitly improves the effectiveness of the target-domain reconstruction loss. Experimental results further validate its superiority over other high-level auxiliary tasks.

    \subsection{Training Loss}

    On the source domain, we train the main task of disparity regression using the per-pixel smooth-\emph{L}1 loss: $L_{s}^{main} = Smooth_{L_{1}}(d_{s}^{l}-\hat{d_{s}^{l}})$. In addition, the per-pixel binary cross entropy loss is adopted for occlusion mask training on the source domain: $L_{s}^{occ} = BCE(O_{s}^{l},\hat{O_{s}^{l}})$.

    On the target domain, the occlusion-aware appearance reconstruction loss is defined as:
    \begin{equation}
    \begin{aligned}
    L_{t}^{ar} = \alpha\frac{1-SSIM(I_{t}^{l}\odot(1-O_{t}^{l}),\overline{I_{t}^{l}}\odot(1-O_{t}^{l}))}{2}  \\
    + \,\,\, (1-\alpha)||I_{t}^{l}\odot(1-O_{t}^{l})-\overline{I_{t}^{l}}\odot(1-O_{t}^{l})||_{1}
    \end{aligned}
    \end{equation}
    where $\odot$ denotes element-wise multiplication, $SSIM$ denotes a simplified single scale SSIM \cite{wang2004image} term with a $3\times3$ block fiter, and $\alpha$ is set to $0.85$. Besides, we apply a \emph{L}1-regularization term on the produced target-domain occlusion mask: $L_{t}^{occ} = ||O_{t}^{l}||_{1}$. Last but not least, we adopt an edge-aware term as the target-domain disparity smoothness loss, where $\partial{I}$ and $\partial{d}$ denote image and disparity gradients:
    \begin{equation}
    L_{t}^{sm} = |\partial_{x}{d_{t}^{l}}|e^{-|\partial_{x}{I_{t}^{l}}|} +  |\partial_{y}{d_{t}^{l}}|e^{-|\partial_{y}{I_{t}^{l}}|}
    \end{equation}

    Finally, the total training loss is a weighted sum of five loss terms mentioned above, where $\lambda_{s}^{occ}$, $\lambda_{t}^{ar}$, $\lambda_{t}^{occ}$, and $\lambda_{t}^{sm}$ denote corresponding loss weights:
    \begin{equation}
    \label{loss_t}
    L = L_{s}^{main} + \lambda_{s}^{occ}L_{s}^{occ} + \lambda_{t}^{ar}L_{t}^{ar} + \lambda_{t}^{occ}L_{t}^{occ} + \lambda_{t}^{sm}L_{t}^{sm}
    \end{equation}

    \begin{table*}[htb]
    \centering
        \caption{Ablation studies on the KITTI, Middleburry, ETH, and DrivingStereo training sets. $D1$-error (\%) is adopted for evaluation.}
        \resizebox{0.99\textwidth}{!}{
        \begin{tabular}{c | c c c | c c | c c | c | c }
	        \hline
            \multirow{2}{*}{Model} & {cost} & {color} & {self-supervised} &
            \multicolumn{2}{c|}{~~~~~~~~KITTI~~~~~~~} &
            \multicolumn{2}{c|}{~~~~Middlebury~~~~} &
            \multirow{2}{*}{~~~ETH~~~} &
            \multirow{2}{*}{~DrivingStereo~} \\
            {} & {normalization} & {transfer} & {reconstruction} &
            {2012} & {2015} & {half} & {quarter} & {} & \\
            \hline
            \multirow{5}{*}{\adapsmnet} &
            {\xmark} & {\xmark} & {\xmark} &
            13.6 & 12.1 & 18.6 & 11.5 & 10.8 & 20.9 \\
            {} & {\cmark} & {\xmark} & {\xmark} &
            11.8 & 9.1 & 16.8 & 10.1 & 9.0 & 16.7 \\
            {} & {\xmark} & {\cmark} & {\xmark} &
            5.3 & 5.4 & 10.0 & 5.8 & 6.1 & 7.4 \\
            {} & {\cmark} & {\cmark} & {\xmark} &
            4.5 & 4.7 & 9.0 & 5.1 & 5.2 & 6.4 \\
            {} & {\cmark} & {\cmark}  & {\cmark} &
            \textbf{3.6} & \textbf{3.5} & \textbf{8.4} & \textbf{4.7} & \textbf{4.1} & \textbf{5.1} \\
            \hline
            \multirow{5}{*}{\adaresnetcorr} &
            {\xmark} & {\xmark} & {\xmark} &
            9.8 & 9.4 & 22.5 & 12.8 & 15.8 & 17.2 \\
            {} & {\cmark} & {\xmark} & {\xmark} &
            8.1 & 8.4 & 19.7 & 10.9 & 13.4 & 15.2 \\
            {} & {\xmark} & {\cmark} & {\xmark} &
            6.7 & 6.7 & 15.1 & 8.3 & 7.1 & 10.2 \\
            {} & {\cmark} & {\cmark} & {\xmark} &
            6.0 & 5.9 & 13.7 & 7.4 & 6.6 & 9.2 \\
            & {\cmark} & {\cmark} & {\cmark} &
            \textbf{5.1} & \textbf{5.0} & \textbf{12.7} & \textbf{6.6} & \textbf{5.8} & \textbf{8.0} \\
	    \hline
        \end{tabular}
        }
    \label{t1}
    \end{table*}

    \section{Experiment}

    To prove the effectiveness of our domain adaptation pipeline, we extend the $2$-D stereo baseline network ResNetCorr \cite{yang2018srcdisp,song2020edgestereo} as \textbf{Ada-ResNetCorr}, and the $3$-D stereo baseline network PSMNet \cite{chang2018pyramid} as \textbf{Ada-PSMNet}. We first conduct detailed ablation studies on multiple datasets including KITTI \cite{Geiger2012CVPR,Menze2015CVPR}, Middlebury \cite{Scharstein2014High}, ETH3D \cite{schoeps2017cvpr}, and DrivingStereo \cite{yang2019drivingstereo}. Next, we compare the cross-domain performance of our domain-adaptive models with other traditional / domain generalization / domain adaptation methods. Finally, we show that our domain-adaptive models achieve remarkable performance on public stereo matching benchmarks.

    \subsection{Datasets}


    The SceneFlow dataset \cite{mayer2016large} is a large synthetic dataset containing $35k$ training pairs with dense ground-truth disparity maps, acting as the source-domain dataset for training.

    The KITTI dataset includes two subsets, \emph{i.e.} KITTI 2012 \cite{Geiger2012CVPR} and KITTI 2015 \cite{Menze2015CVPR}, providing $394$ stereo pairs of outdoor driving scenes with sparse ground-truth disparities for training, and $395$ pairs for testing. The Middlebury dataset \cite{Scharstein2014High} is a small indoor dataset containing less than $50$ stereo pairs with three different resolutions. The ETH3D dataset \cite{schoeps2017cvpr} includes both indoor and outdoor scenarios, containing $27$ gray-image pairs with dense ground-truth disparities for training, and $20$ pairs for testing. The DrivingStereo dataset \cite{yang2019drivingstereo} is a large-scale stereo matching dataset covering a diverse set of driving scenarios, containing over $170k$ stereo pairs for training and $7751$ pairs for testing. These four real-world datasets act as different target domains that are adopted for cross-domain evaluations.

    We adopt the bad pixel error rate ($D1$-error) as the evaluation metric, which calculates the percentage of pixels whose disparity errors are greater than a certain threshold.

    \subsection{Implementation Details}

    Each model is trained end-to-end using the Adam optimizer ($\beta_1=0.9$, $\beta_2=0.999$) on eight NVIDIA Tesla-V100 GPUs. The learning rate is set to $0.001$ for training from scratch, and we train each model for $100$ epochs with a batch size of $16$ using $624\times304$ random crops. The momentum factor $\gamma$ in Alg. \ref{alg:color_transfer} is set to $0.95$. The weights of different loss terms ($\lambda_{s}^{occ}$, $\lambda_{t}^{ar}$, $\lambda_{t}^{occ}$,  $\lambda_{t}^{sm}$) in Eq.~\ref{loss_t} are set to ($0.2$, $1.0$, $0.2$, $0.1$). An individual domain-adaptive stereo model is trained for each target domain. The model specifications of the Ada-ResNetCorr and Ada-PSMNet are provided in the supplementary material.

   \subsection{Ablation Studies}
   \label{subsec:ablation}

   In Tab. \ref{t1}, we conduct detailed ablation studies on four real-world datasets to evaluate the key components in our domain adaptation pipeline, based on Ada-PSMNet and Ada-ResNetCorr. As can be seen, applying the progressive color transfer algorithm during  training can significantly reduce error rates on multiple target domains, \emph{e.g.} $8.3\%$ on KITTI, $8.6\%$ on Middlebury, $4.7\%$ on ETH3D, and $13.5\%$ on DrivingStereo from Ada-PSMNet, benefiting from massive color-aligned training images without geometrical distortions. We also provide qualitative results of color transfer in the supplementary material. In addition, compared with baseline models, error rates are reduced by $1\%\sim4\%$ on varied target domains by integrating the proposed cost normalization layer, which also works well when implemented together with the input color transfer module. Furthermore, adopting the self-supervised occlusion-aware reconstruction can further reduce error rates by $0.5\%\sim1.5\%$ on varied target domains, though the adaptation performance is already remarkable through color transfer and cost normalization. Finally, both Ada-PSMNet and Ada-ResNetCorr significantly outperform the corresponding baseline model on all target domains, especially an accuracy improvement of $15.8\%$ from Ada-PSMNet on the large-scale DrivingStereo dataset.

   In order to further demonstrate the effectiveness of each module, we perform exhaustive comparisons with other alternative methods respectively. As shown in Tab.~\ref{t6}, our specifically designed cost normalization layer which is parameter-free and adopted only once in the network, outperforms other general and learnable normalization layers (AdaBN \cite{li2016revisiting}, BIN \cite{nam2018batch}, IN \cite{ulyanov2016instance}, and DN \cite{zhang2020domain}) which are repeatedly adopted in the network's feature extractor. In Tab.  \ref{t7}, our progressive color transfer algorithm far outperforms three popular color/style transfer networks  (WCT$^{2}$~\cite{yoo2019photorealistic}, WaterGAN \cite{li2017watergan}, and CycleGAN \cite{zhu2017unpaired}), indicating that geometrical distortions from such GAN-based color/style transfer models are harmful for the low-level stereo matching task. Moreover, our method outperforms the Reinhard's color transfer method \cite{reinhard2001color} by about $1\%$ in $D1$-error, revealing the effectiveness of the proposed progressive update strategy. In Tab.  \ref{t8}, our proposed self-supervised occlusion-aware reconstruction is demonstrated to be a more effective auxiliary task for stereo domain adaptation, while other alternatives all hurt the domain adaptation performance.


    \begin{table}[t]
    \centering
    \caption{
        Comparisons with existing normalization layers on the KITTI and DrivingStereo training sets. $D1$-error (\%) is adopted.
    }
    \resizebox{0.9\linewidth}{!}{
        \begin{tabular}{c | c  c}
	        \hline
            {Methods} &
            {KITTI} &
            {DrivingStereo} \\
            \hline
            {PSMNet Baseline}    & 12.1 & 20.9 \\
            {+Adaptive Batch Norm \cite{li2016revisiting}}    & 11.8 & 20.3 \\
            {+Batch-Instance Norm \cite{nam2018batch}}    & 11.2 & 19.5 \\
            {+Instance Norm \cite{ulyanov2016instance}} & 10.7 & 18.6 \\
            {+Domain Norm \cite{zhang2020domain}} & 9.5 & 17.2 \\
            {+Our Cost Norm} & \textbf{9.1} & \textbf{16.7} \\
            \hline
        \end{tabular}
    }
    \label{t6}
    \end{table}

    \begin{table}[t]
    \centering
    \caption{
        Comparisons with color/style transfer methods on the KITTI and DrivingStereo training sets. $D1$-error (\%) is adopted.
    }
    \resizebox{0.99\linewidth}{!}{
        \begin{tabular}{c | c  c}
	        \hline
            {Methods} &
            {KITTI} &
            {DrivingStereo} \\
            \hline
            {PSMNet Baseline}    & 12.1 & 20.9 \\
            {+WCT$^{2}$ \cite{yoo2019photorealistic}} & 10.2 & 17.3 \\
            {+WaterGAN \cite{li2017watergan}} & 8.7 & 11.5 \\
            {+CycleGAN \cite{zhu2017unpaired}} & 8.0 & 10.6 \\
            {+Color Transfer \cite{reinhard2001color}} & 6.2 & 8.3 \\
            {+Our Progressive Color Transfer} & \textbf{5.4} & \textbf{7.4} \\
            \hline
        \end{tabular}
    }
    \label{t7}
    \end{table}

    \begin{table}[t]
    \centering
    \caption{
        Comparisons with other auxiliary tasks for stereo domain adaptation on the KITTI training set. $D1$-error (\%) is adopted.
    }
    \resizebox{1\linewidth}{!}{
        \begin{tabular}{c | c}
	        \hline
	        Methods & KITTI  \\
	        \hline
	        PSMNet + Our Color Transfer + Our Cost Norm (Baseline) & 4.7 \\
	        +Patch Localization Task \cite{xu2019self} & 6.5 \\
	        +Rotation Prediction Task \cite{gidaris2018unsupervised} & 6.1 \\
	        +Flip Prediction Task \cite{xu2019self} & 6.0  \\
	        +Our Self-Supervised Reconstruction Task & \textbf{3.5} \\
	        \hline
        \end{tabular}
    }
    \label{t8}
    \end{table}

    \subsection{Cross-Domain Comparisons}

\begin{figure}[tb]
    	\centering
    	\includegraphics[width=0.97\linewidth]{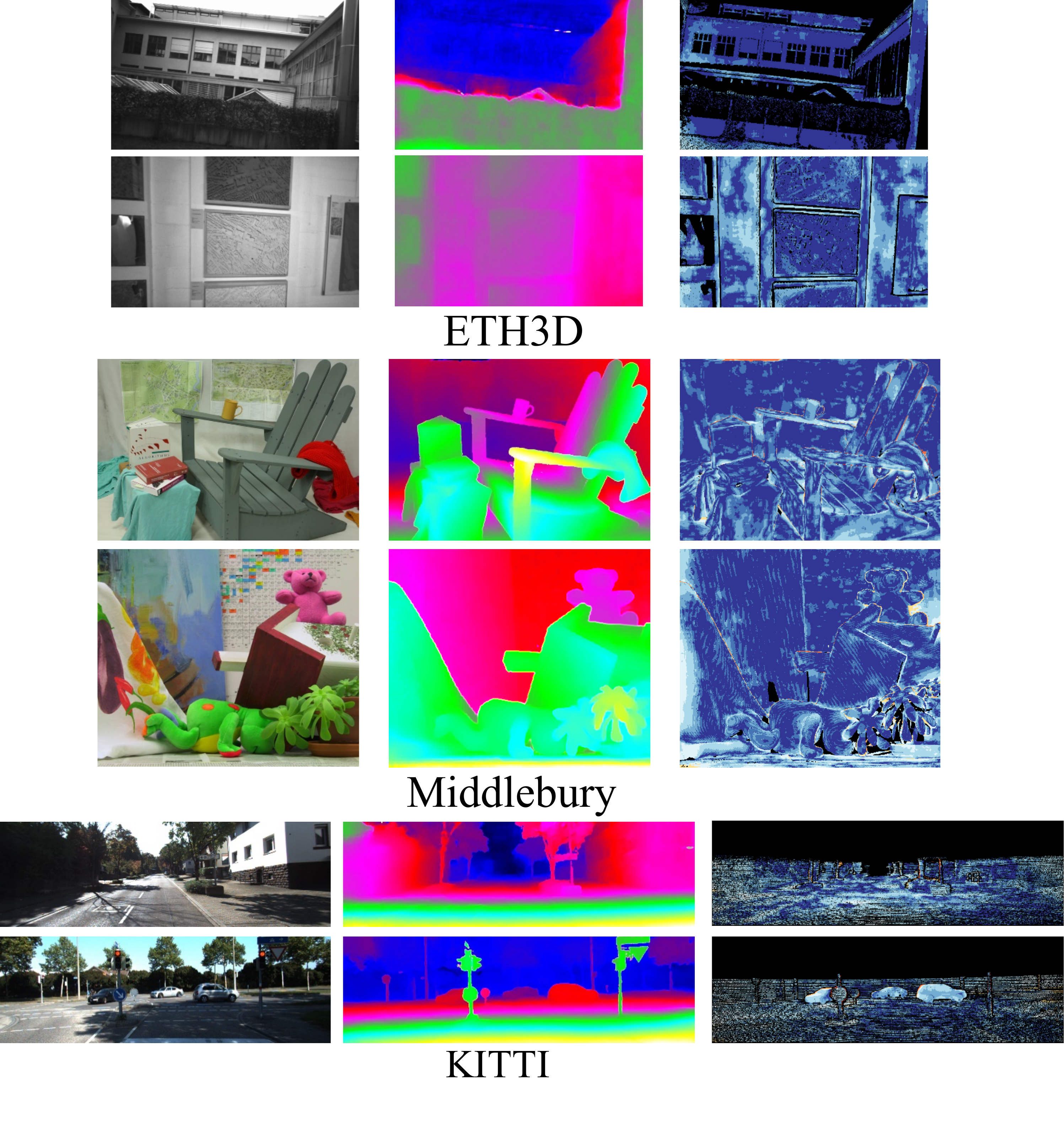}
    	\caption{Disparity predictions from our SceneFlow-pretrained Ada-PSMNet on the ETH3D, Middlebury, and KITTI datasets. Left-right: left image, colorized disparity  map, and error map.
    	}
    \label{fig:disp_pred}
    \end{figure}

    In Tab. \ref{t2}, we compare our proposed domain-adaptive stereo models with other traditional stereo methods, domain generalization, and domain adaptation stereo networks on three real-world datasets. Firstly, both Ada-ResNetCorr and Ada-PSMNet show great superiority over traditional stereo methods. Secondly, for comparisons with domain generalization networks, unfairness may exist since our domain-adaptive models use target-domain images during training. It is caused by the problem definition of domain adaptation as mentioned in Sec. \ref{sec:formulation}. However, as can be seen in Tab.~\ref{t2}, our Ada-PSMNet achieves tremendous gains rather than small deltas compared with all domain generalization networks, including the state-of-the-art DSMNet \cite{zhang2020domain} and its baseline network GANet \cite{zhang2019ga}. Lastly, among the few published domain adaptation networks, only the StereoGAN \cite{liu2020stereogan} reported such cross-domain performance, while our Ada-PSMNet achieves a $3.5$ times lower error rate than StereoGAN~\cite{liu2020stereogan} on the KITTI training set. Hence, our proposed multi-level alignment pipeline successfully address the domain adaptation problem for stereo matching. In Fig. \ref{fig:disp_pred}, we provide qualitative results of our method on different real-world datasets, in which Ada-PSMNet predicts accurate disparity maps on both outdoor and indoor scenes.

    \begin{table}[t]
    \centering
    \caption{
        Cross-domain comparisons with other traditional / domain generalization / domain adaptation stereo methods on the KITTI, Middlebury, and ETH3D training sets. $D1$-error (\%) is adopted. The second and third columns indicate whether the method is trained on the SceneFlow dataset, and whether the method uses target-domain images during training respectively.
    }
    \resizebox{1\linewidth}{!}{
        \begin{tabular}{c | c | c | c  c  c}
	        \hline
            \multirow{2}{*}{Methods} &
            {Train} &
            {Target} &
            \multicolumn{3}{c}{Test} \\
            {} &
            {SceneFlow} &
            {Images} &
            {~~KITTI~~} &
            {Middlebury} &
            {~~ETH~~} \\
            \hline
            \multicolumn{6}{c}{Traditional Stereo Methods} \\
            \hline
            {PatchMatch \cite{bleyer2011patchmatch}}    & {\xmark} & {\xmark} & 17.2 & 38.6 & 24.1 \\
            {SGM \cite{hirschmuller2008stereo}}           & {\xmark} & {\xmark} & 7.6 & 25.2  & 12.9 \\
            \hline
            \multicolumn{6}{c}{Domain Generalization Stereo Networks} \\
            \hline
            {HD$^{3}$ \cite{yin2019hierarchical}}      & {\cmark} & {\xmark} & 26.5  & 37.9  & 54.2 \\
            {GWCNet \cite{guo2019group}}        & {\cmark} & {\xmark} & 22.7  & 34.2  & 30.1 \\
            {PSMNet \cite{chang2018pyramid}}        & {\cmark} & {\xmark} & 16.3  & 25.1  & 23.8 \\
            {GANet \cite{zhang2019ga}}         & {\cmark} & {\xmark} & 11.7  & 20.3  & 14.1 \\
            {DSMNet \cite{zhang2020domain}}        & {\cmark} & {\xmark} & 6.5   & 13.8  & 6.2 \\
            \hline
            \multicolumn{6}{c}{Domain Adaptation Stereo Networks} \\
            \hline
            {StereoGAN \cite{liu2020stereogan}}      & {\cmark} & {\cmark} & 12.1  & -     & - \\
            {\adaresnetcorr} & {\cmark} & {\cmark} & 5.0 & 12.7 & 5.8 \\
            {\adapsmnet}     & {\cmark} & {\cmark} & \textbf{3.5} & \textbf{8.4} & \textbf{4.1} \\
            \hline
        \end{tabular}
    }
    \label{t2}
    \end{table}

    \begin{table*}[!t]
    \centering
        \caption{Performance on the ETH3D stereo benchmark. The $1$-pixel error (\%) and $2$-pixel error (\%) are adopted for evaluation.}
        \resizebox{0.9\textwidth}{!}{
            \begin{tabular}{ c | c | c | c | c | c | c | c }
            \hline
            {Method} & {Deep-Pruner} & {iResNet} & {Stereo-DRNet}  & {SGM-Forest} & {PSMNet} &  {DispNet} & {\adapsmnet}\\
            {Use ETH-gt} & {\cmark} & {\cmark} & {\cmark} & {\xmark} & {\cmark} & {\cmark} & {\xmark} \\
            \hline
            {Bad 1.0} & 3.52 & 3.68 & 4.46 & 4.96 & 5.02 & 17.47 & \textbf{3.09} \\
            {Bad 2.0} & 0.86 & 1.00 & 0.83 & 1.84 & 1.09 & 7.91  & \textbf{0.65} \\
            \hline
            \end{tabular}
        }
    \label{t3}
    \end{table*}

    \begin{table*}[!t]
    \centering
        \caption{Performance on the Middlebury stereo benchmark. The  $2$-pixel error (\%) is adopted for evaluation.}
        \resizebox{0.9\textwidth}{!}{
            \begin{tabular}{ c | c | c | c | c | c | c | c }
            \hline
            {Method} & {EdgeStereo} & {CasStereo} & {iResNet} & {MCV-MFC} & {Deep-Pruner} & {PSMNet} & {\adapsmnet}\\
            {Use Mid-gt} & {\cmark} & {\cmark} & {\cmark} & {\cmark} & {\cmark} & {\cmark} & {\xmark} \\
            \hline
            Bad 2.0 & 18.7 & 18.8 & 22.9 & 24.8 & 30.1 & 42.1 & \textbf{13.7} \\
            \hline
            \end{tabular}
        }
    \label{t4}
    \end{table*}

	\begin{table*}[!t]
    \centering
        \caption{Performance on the KITTI 2015 stereo benchmark. The $D1$-error (\%) is adopted for evaluation.}
        \resizebox{0.99\textwidth}{!}{
            \begin{tabular}{ c | c | c | c | c | c | c | c |c |c}
            \hline
            {Method} & {GC-Net} & {L-ResMatch} & {SGM-Net} & {MC-CNN} & {DispNetC}  & {Weak-Sup} & {MADNet} & {Unsupervised} & {\adapsmnet} \\
            {Use KITTI-gt} & {\cmark} & {\cmark} & {\cmark} & {\cmark} & {\cmark} & {\cmark} & {\xmark} & {\xmark} & {\xmark} \\
            \hline
            $D1$-error & \textbf{2.87} & 3.42 & 3.66 & 3.89 & 4.34 & 4.97 & 8.23 & 9.91 & 3.08 \\
            \hline
            \end{tabular}
        }
    \label{t5}
    \end{table*}

    \subsection{Evaluations on Stereo Benchmarks}

    We further compare our domain-adaptive stereo model Ada-PSMNet with several unsupervised/self-supervised methods and finetuned disparity networks on public stereo matching benchmarks: KITTI, Middlebury, and ETH3D. We directly upload the results from our SceneFlow-pretrained model to the online benchmark, and do not finetune using target-domain ground-truths before submitting test results.

	\subsubsection{Results on the ETH3D Benchmark}
	
    As can be seen in Tab. \ref{t3}, SceneFlow-pretrained Ada-PSMNet outperforms the state-of-the-art patch-based model DeepPruner \cite{duggal2019deeppruner}, end-to-end disparity networks (iResNet \cite{liang2017learning}, PSMNet \cite{chang2018pyramid}, and StereoDRNet \cite{chabra2019stereodrnet}) finetuned with ground-truth disparities from the ETH3D training set, and state-of-the-art traditional method SGM-Forest \cite{schonberger2018learning}. By the time of the paper submission, AdaStereo ranks 1$^{st}$ on the ETH3D benchmark in terms of the $2$-pixel error metric.

    \subsubsection{Results on the Middlebury Benchmark}

    As can be seen in Tab.~\ref{t4}, SceneFlow-pretrained Ada-PSMNet significantly  outperforms all other state-of-the-art end-to-end disparity networks (EdgeStereo \cite{song2020edgestereo}, CasStereo \cite{gu2020cascade}, iResNet \cite{liang2017learning}, MCV-MFC \cite{liang2019stereo}, and PSMNet \cite{chang2018pyramid}) which are finetuned using ground-truth disparities from the Middlebury training set. Our Ada-PSMNet achieves a remarkable $2$-pixel error rate of $13.7\%$ on the full-resolution test set, outperforming all other finetuned end-to-end stereo matching networks on the benchmark.

    \subsubsection{Results on the KITTI Benchmark}

    As can be seen in Tab.  \ref{t5}, SceneFlow-pretrained Ada-PSMNet far outperforms the online-adaptive model MADNet \cite{tonioni2019real}, weakly supervised \cite{tulyakov2017weakly} and unsupervised \cite{zhou2017unsupervised} methods, meanwhile achieving higher accuracy than some supervised disparity networks including  MC-CNN-acrt \cite{zbontar2015computing},  L-ResMatch \cite{shaked2017improved}, DispNetC \cite{mayer2016large}, and SGM-Net \cite{seki2017sgm}. Moreover, our  Ada-PSMNet achieves comparable performance with the KITTI-finetuned GC-Net \cite{kendall2017end}.

\section{Conclusions}

In this paper, we focus on the domain adaptation problem for deep stereo networks. Following the standard domain adaptation methodology, we propose a novel domain adaptation pipeline specifically for stereo matching task, in which multi-level alignments are conducted: a non-adversarial progressive color transfer algorithm for input-level alignment; a parameter-free cost normalization layer for internal feature-level alignment; a highly related self-supervised auxiliary task for output-space alignment. We verify our SceneFlow-pretrained domain-adaptive models on four real-world datasets, and state-of-the-art cross-domain performance is achieved on all target domains. Our AdaStereo  model also achieves remarkable performance on multiple stereo matching benchmarks without finetuning.

\bibliographystyle{ieee_fullname}
\bibliography{egbib}

\end{document}